\documentclass[10pt,twocolumn,letterpaper]{article}

\usepackage[pagenumbers]{cvpr} 
\usepackage{tabularx}
\usepackage{makecell}
\usepackage{multirow}
\usepackage[dvipsnames,table]{xcolor}

\definecolor{danred}{rgb}{0.9098,0.9098,0.9098}

\usepackage{tablefootnote}
\definecolor{cvprblue}{rgb}{0.21,0.49,0.74}

\usepackage[pagebackref,breaklinks,colorlinks,citecolor=cvprblue]{hyperref}

\title{DreamVideo: High-Fidelity Image-to-Video Generation \\with Image Retention and Text Guidance}

\author{Cong~Wang$^{1}$\quad Jiaxi~Gu$^{2}$\quad Panwen~Hu$^{3}$\quad Songcen~Xu$^{2}$\quad Hang~Xu$^{2}$\quad Xiaodan~Liang $^{1}$\footnotemark[1] \\
$^{1}$ Shenzhen Campus of Sun Yat-Sen University, China \\
$^{2}$Huawei Noah’s Ark Lab, China  \quad $^{3}$
Shenzhen Campus of the Chinese University of Hong Kong\\
{\tt\small $^{1}$\{wangc39@mail2., liangxd9@mail.\}sysu.edu.cn\quad} \\ {\tt\small $^{2}$\{imjiaxi, chromexbjxh\}@gmail.com} {\tt\small $^{3}$\{panwenhu\}@link.cuhk.edu.cn}
}

\begin{document}

\twocolumn[{
\maketitle
\renewcommand\twocolumn[1][]{#1}
    \begin{center}
        \centering
        \includegraphics[width=0.92\linewidth]{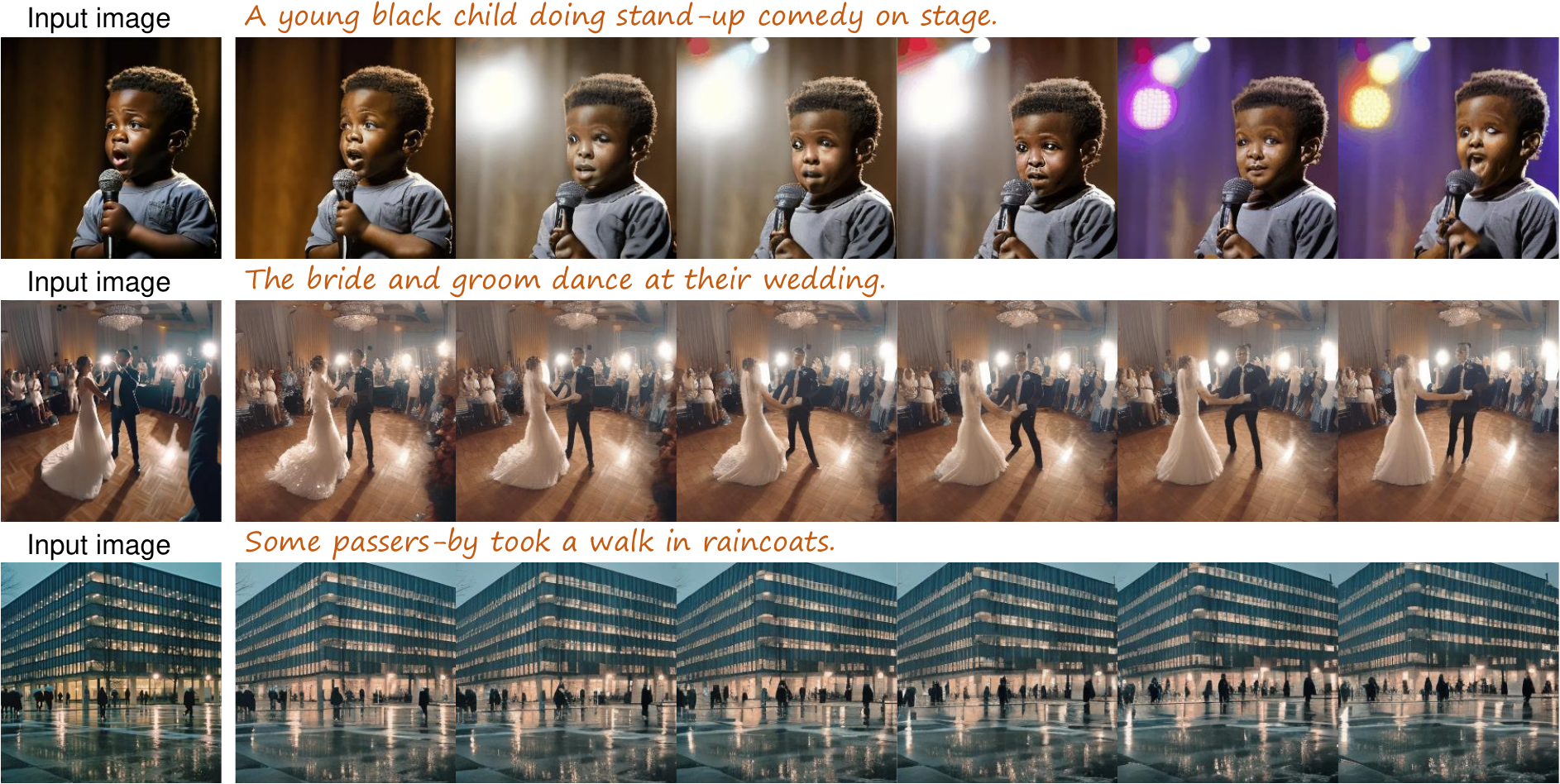}
        \vspace{-3mm}\captionof{figure}{With DreamVideos, high fidelity is obtained between the input image and the first frame of the generated video, e.g., ``a young black child''. Moreover, text guidance also helps to control the motion of the video content including ``dance'' and ``walk''. The images in showcases are from MidJourney.}
        \label{fig:showcase}
        \vspace{8mm}
    \end{center}
}]

\renewcommand{\thefootnote}{\fnsymbol{footnote}} 
\footnotetext[1]{Corresponding authors.} 

\begin{abstract}

Image-to-video generation, which aims to generate a video starting from a given reference image, has drawn great attention. Existing methods try to extend pre-trained text-guided image diffusion models to image-guided video generation models. Nevertheless, these methods often result in either low fidelity or flickering over time due to their limitation to shallow image guidance and poor temporal consistency. To tackle these problems, we propose a high-fidelity image-to-video generation method by devising a frame retention branch based on a pre-trained video diffusion model, named \textbf{DreamVideo}. Instead of integrating the reference image into the diffusion process at a semantic level, our DreamVideo perceives the reference image via convolution layers and concatenates the features with the noisy latents as model input. By this means, the details of the reference image can be preserved to the greatest extent. In addition, by incorporating double-condition classifier-free guidance, a single image can be directed to videos of different actions by providing varying prompt texts. This has significant implications for controllable video generation and holds broad application prospects. We conduct comprehensive experiments on the public dataset, and both quantitative and qualitative results indicate that our method outperforms the state-of-the-art method.  Especially for fidelity, our model has a powerful image retention ability and delivers the best results in UCF101 compared to other image-to-video models to our best knowledge. Also, precise control can be achieved by giving different text prompts. Further details and comprehensive results of our model will be presented in \url{https://anonymous0769.github.io/DreamVideo/}.


\end{abstract}

    
\section{Introduction}
\label{sec:intro}

Due to the large amount of potential applications in artistic creation, entertainment, etc., image or video generation has attracted much attention. Text-guided image generation has experienced explosive growth and a large number of excellent works have been published such as GLIDE~\cite{saharia2022photorealistic}, Imagen~\cite{ho2022imagen}, Stable Diffusion~\cite{rombach2022high}, et al. Leveraging existing text-guided image synthesis technologies, text-to-video generation~\cite{singer2022make} has also achieved certain milestones. Text-to-video generation can be seen as a simple extension of text-to-image in the video domain by injecting temporal layers into the network with spatial layers frozen. The ability of text-guided video generation is however limited in real applications. The reasons are twofold. One reason is that a simple and short sentence can barely describe a video containing dynamic content. The other reason is that complex texts present a significant challenge to the representational capabilities of the generation model, thereby greatly increasing the likelihood of generation failure. For this purpose, various methods have been proposed for generating videos guided by different conditions including optical flows~\cite{ni2023conditional}, depth sequence~\cite{wang2023videocomposer}, dragging strokes~\cite{yin2023dragnuwa}, et al. Among all the conditions, the condition of image is a natural and intuitive way to guide video generation. On the one hand, an image provides substantial visual details. On the other hand, text prompts can still play a role in controlling the diversity of video generation. Therefore, image-to-video generation with text guidance holds significant importance.

Image-to-video generation with text guidance is a straightforward idea but achieving high quality is nontrivial. The biggest challenge is how to enable flexible control over various video content with text guidance while maintaining the fidelity to the input image. Usually, visual details are sacrificed for temporal consistency across generated video frames, which leads to flickering over the generated video frames. For the same reason, all of the generated video frames differ from the input one so the fidelity gets low. Moreover, even if both fidelity and temporal consistency can be maintained, the problem of the limited capability of video motion control by text guidance, as another challenge, arises. Previous works have attempted to process the input image as an additional condition to text prompt~\cite{chen2023videocrafter1}. The result is unsatisfactory especially in fidelity because visual information is only captured at the semantic level. To tackle this problem, there are also works~\cite{zhang2023i2vgen} using two-stage training on images and then texts respectively by using cascaded diffusion models. The authors try to keep visual fidelity to the input image and meanwhile hold a good video-text alignment via two-stage training. Such a complex design partially alleviates the problem of image-to-video generation we previously mentioned. However, the later refinement stage for text-guided video generation plays a greater role, even weakening the effects of the first image-guided stage. The result of this structure is that there is a challenging contradiction between fidelity and text alignment. This issue is also evident in the generation results of the I2VGen-XL~\cite{zhang2023i2vgen}.

To tackle these problems, we propose a high-fidelity image-to-video generation method by devising a frame retention branch on the basis of a pre-trained video diffusion model. Instead of integrating the reference image into the diffusion process at a semantic level, our method perceives the reference image via convolution layers and concatenates the features with the noisy latents as model input. By this means, the details of the reference image can be preserved to the greatest extent. In addition, by incorporating double-condition classifier-free guidance, a single image can be directed to videos of different actions by providing varying prompt texts. \Cref{fig:showcase} exhibits three cases of image-to-video generation using DreamVideo, with the image examples sourced from MidJourney~\cite{MidJourney}, which have high fidelity and great text alignment. Hence, our DreamVideo can seamlessly integrate with the text-to-image generation model, thereby facilitating a production pipeline from text to image to video.
Moreover, to fully evaluate our proposed method, extensive experiments are conducted on multiple benchmarks. On the benchmark of image-video generation based on UCF101, we achieve a considerable Fréchet Video Distance (FVD)~\cite{unterthiner2019fvd} score of 214.52, which significantly outperforms most existing methods. Also, the comparisons with other methods are conducted from multiple aspects including user study.

The contributions of our work can be summarized as follows:
\begin{itemize}
    \item We devise a high-fidelity image-to-video generation model through a carefully designed and cheap-to-train frame retention module.
    \item We achieve flexible motion control for animating a given image while maintaining high fidelity.
    \item Extensive experiments show that our proposed method can achieve state-of-the-art in UCF101 and MSRVTT datasets and show better image retention and video generation quality in comparison with other methods.
    
\end{itemize}

\section{Related work}
\label{sec:rel}

\subsection{Video diffusion models}

Diffusion Models (DMs)~\cite{2020DMs} show impressive results in image synthesis and lots of methods have been proposed such as GLIDE~\cite{nichol2021glide}, Imagen~\cite{saharia2022photorealistic}, Stable Diffusion~\cite{rombach2022high}, et al. With the maturity of image generation models, video generation also receives much attention and the widely adopted approach is injecting temporal layers into the image DMs to enable the network temporal representation. Lots of video diffusion models have been published such as Make-A-Video~\cite{singer2022make}, CogVideo~\cite{hong2022cogvideo}, Imagen Video~\cite{ho2022imagen}, MagicVideo~\cite{zhou2022magicvideo}, VidRD~\cite{gu2023reuse}, et al. For the training data, many works~\cite{blattmann2023align,wang2023lavie} find that using both image and video data can significantly improve the appearance details and also alleviate the phenomenon of catastrophic forgetting. For the complexity of video data, PVDM~\cite{yu2023video} creatively proposes an image-like 2D latent space for efficient parameterization. Also, some works study the effects of initial noise prior in video DMs. VideoFusion~\cite{luo2023videofusion} finds the image priors of the pre-trained model can be efficiently shared by all frames and thereby facilitate the learning of video data. PYoCo~\cite{ge2023preserve} devises a video noise prior to getting better temporal consistency. In addition, some works use additional DMs for frame interpolation, prediction and super resolution to achieve better performance. Align Your Latent~\cite{blattmann2023align} and LAVIE~\cite{wang2023lavie} are two relatively complete pipelines to generate high-quality videos. Both of them devise a basic video DM to generate the initial video frames and additional modules for temporal interpolation and Video Super Resolution (VSR).

\subsection{Image-to-video generation}

Earlier video generation works following image diffusion models are mostly text-guided. Though creative videos can indeed be generated with carefully devised text prompts, the appearance, layout or motion of the generated videos can not be precisely controlled. For this purpose, lots of works try to integrate other conditions or controls into video DMs. Image-to-video generation is a straightforward idea but not a trivial one to implement. AnimateDiff~\cite{guo2023animatediff} injects motion modeling modules into frozen text-to-image models so motion priors can be distilled for different personalized models to implement image-to-video generation. VideoComposer~\cite{wang2023videocomposer} devises a unified interface for multiple conditions including single image condition. Also, there are some works exploiting additional modalities to generate videos from images. Dreampose~\cite{karras2023dreampose} adopts a sequence of poses and LFDM~\cite{ni2023conditional} uses optical flows. Additionally, DragNUWA~\cite{yin2023dragnuwa} pushes image-to-video ahead by integrating user strokes to realize more controllable video generation. VideoCrafter1~\cite{chen2023videocrafter1} treats the input image as an additional condition and devises a dual attention module to encode and inject both text prompt and reference image. I2VGen-XL~\cite{zhang2023i2vgen} uses two stages of cascaded diffusion models to achieve high semantic consistency and spatiotemporal continuity. However, most of these works on image-to-video generation are semantic level so the details of objects often differ from those in the input image. Therefore a high-fidelity image-to-video generation method is necessary.

\section{Preliminary}

\textbf{Diffusion models (DMs)~\cite{2020DMs}} is a probabilistic generative model that learns the underlying data distribution through two steps: diffusion and denoising. Specifically, during the diffusion process, given an input data $\mathbf{z}$, the model gradually adds random noise $ \mathbf{z}_{t} = \alpha_{t}\mathbf{z} + \sigma_{t}\epsilon $, where $ \epsilon \in \mathcal{N}(\mathbf{0}, \mathbf{I}) $. The magnitudes of noise addition are controlled by $ \alpha_{t} $ and $ \sigma_{t} $ as the denoising steps $t$ progress. In the following denoising stage, the model takes the diffused sample $ \mathbf{z}_{t} $ as input and minimizes the mean squared error loss to learn a denoising function $ \epsilon_{\theta} $ as follows: 
\begin{equation}
    E_{\mathbf{z},\epsilon,t}=\left[ \lVert \epsilon_{\theta}(\mathbf{z}_{t}, t) - \epsilon \lVert \right]
\end{equation}

\noindent \textbf{Latent Diffusion Models (LDMs)~\cite{LDMs}} utilize the architecture of Variational Autoencoders (VAEs). Unlike Diffusion Models (DMs), LDMs can compress the input data into a latent variable $ \epsilon(\mathbf{z}) $ by encoder $ \epsilon $, and then perform denoising truncation by decoding $D(\mathbf{z}_{0})$ by the decoder $D$. LDMs significantly reduce the training and inference time as the diffusion and denoising are performed in the latent space rather than the data space. The objective of LDMs can be formulated as follows:
\begin{equation}
    E_{\mathbf{z},\epsilon,t}=\left[ \lVert \epsilon_{\theta}(\mathcal{E}(\mathbf{z}_{t}), t) - \epsilon \lVert \right]
\end{equation}

\noindent \textbf{Video Latent Diffusion Models (VLDMs)} build upon LDMs by incorporating a temporal module to capture the temporal continuity in video data. VLDMs typically add a temporal attention module to the U-Net architecture, enabling attention in the temporal dimension. Additionally, the 2d convolutions are modified to 3d convolutions to accommodate video data. Given the impressive capabilities of VLDMs in video generation, our proposed approach, dragVideo, follows the principles of VLDMs, which encodes video into latent variables $ \mathcal{E}(\mathbf{z}) $ and leverages U-Net to learn the spatio-temporal characteristics of the video data.

\section{Method}
\label{sec:method}
The model we propose, DreamVideo, is a novel video synthesis model with text and image as control conditions. The overall architecture of DreamVideo is shown in \cref{fig:model}, consisting of two different networks: a primary Text-to-Video (T2V) model tasked with video generation corresponding to textual inputs, and an Image Retention block that systematically infuses image control signals extracted from the input image into the U-Net~\cite{U-Net} structure. Consequently, DreamVideo possesses the capability to generate tailored videos by harmoniously merging two varying types of input signals, thereby enabling support for a diverse array of downstream tasks. The forthcoming sections will elaborate on DreamVideo's model design, elucidate the integration of control signals into its network, and illustrate how inference is realized under given conditions.

\subsection{Model Structure}
\label{model_structure}

\begin{figure*}[!ht]
    \centering
    \includegraphics[width=1.0\linewidth]{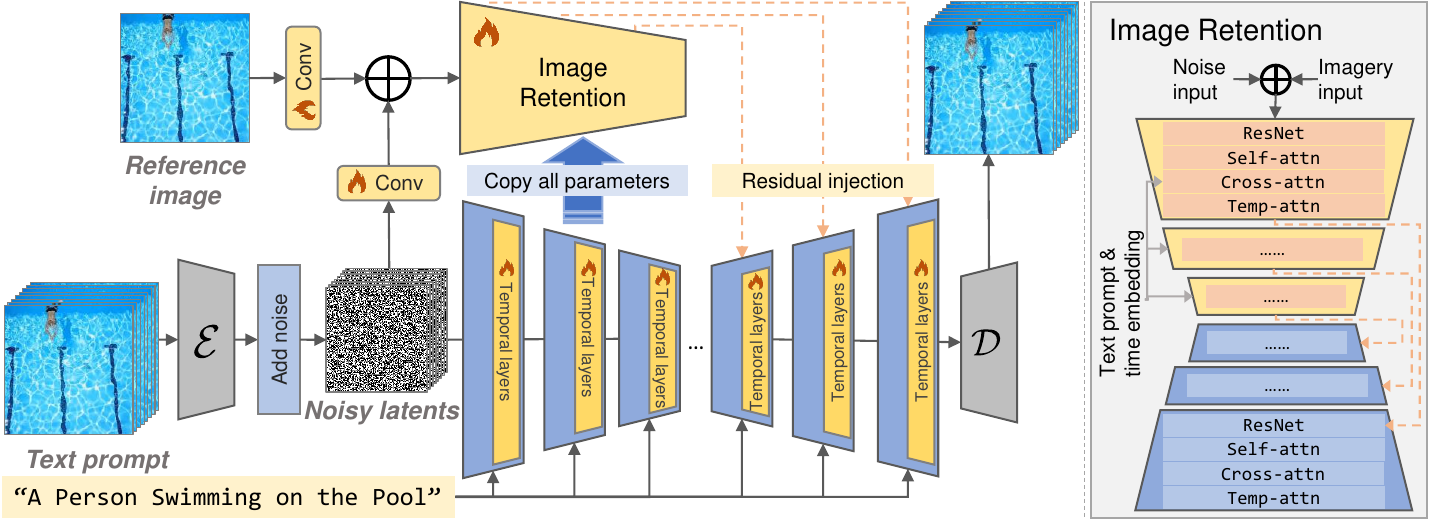}
    \caption{The architecture of DreamVideo. A reference image is processed by a convolution block and concatenated with the representation of noisy latents. The Image Retention Module, as a side branch copying from the downsample blocks of U-Net, plays a role in maintaining the visual details from the input image and meanwhile also accepting text prompts for motion control.}
    \label{fig:model}
\end{figure*}
 
The architecture of our model is shown in \cref{fig:model}, which comprises two principal modules, the primary U-Net block and the Image Retention block. The primary U-Net block forms the foundation of VLDMs and is comprised of a Variational Auto-Encoder (VAE)~\cite{VAE} and U-Net. The VAE takes on the responsibility of encoding and decoding latent representations, whereas U-Net effectively handles denoising. The disparity between our U-Net and the one in Stable Diffusion~\cite{LDMs} lies in the addition of Temp-Attn and Temp-Conv within each sample block in our model. These elements play a pivotal role in preserving the temporal coherence of the generated video.
The Image Retention block encompasses the duplication of network parameters in every down-sample and middle layer of the primary U-Net. In addition, zero-convolutional layers are added at the end of each block to progressively increase the image retention control over the U-Net, which also can preserve the text-video generation ability of the primary U-Net. To facilitate the input of the video latent representation, the 2D convolutions within the network have been modified to 3D convolutions.

Given video data $ \mathbf{x} \in \mathbb{R}^{T \times H \times W \times 3} $, where $T$, $H$, and $W$ denote the number of frames, height, and width of the video in pixel space, respectively. The video data $ \mathbf{x} $ projects into a noisy latent representation $ \mathbf{z}=\mathcal{E}(\mathbf{x}) $ by the encode in VAE, where $ \mathbf{z} \in \mathbb{R}^{T \times c \times h \times w  } $, with $c$, $h$, and $w$ denote the latent channel, height, and width in the latent space, respectively. 
For a provided input control image $ I \in \mathbb{R}^{H \times W \times 3} $, the Image Retention Block leverages convolutional layers to map $ I $ onto the latent space $ \mathbf{z}_{s} \in h \in \mathbb{R}^{c_{I} \times h_{I} \times w_{I}} $. The control signal output gathered from the Image Retention Block, which is derived by integrating the noisy $ \mathbf{z} $ and image latents $ \mathbf{z}_{s} $, is added incrementally to the U-Net.  Eventually, a decoder $ \mathcal{D} $ is adopted to convert the latent variables passing through the up-sample block back to the pixel space.

\subsection{Image Retention}\label{image-retention}

In this section, we elucidate the structure, inputs, and outputs of the Image Retention block, along with its mechanism in how to achieve image retention. The architecture of this module is depicted on the right side of \cref{fig:model}. Within the U-Net framework, we categorize the mid-sample, down-sample, and up-sample blocks as \( F^{m} \), \( F_{i}^{d} \), and \( F_{i}^{u} \) respectively, where \( i \) signifies the layer index. The Image Retention block is structured analogously to the mid-sample and down-sample blocks in U-Net, encompassing components such as ResNet, self-attention, cross-attention, and temporal-attention, represented by \( R_{i} \).

The module accepts dual inputs. The first is the latent representation \( \mathbf{z}_{d} \), which is derived from the output of the convolutional layer: \( \mathbf{z}_{d} = \text{conv}_{d}(\mathbf{z}) \in \mathbb{R}^{T \times c_{0} \times h_{0} \times w_{0}} \). The second input is the image latent representation \( \mathbf{z}_{s} \), which is obtained by processing the image \( I \) through a separate convolutional layer: \( \mathbf{z}_{s} = \text{conv}_{s}(I) \in \mathbb{R}^{1 \times c_{0} \times h_{0} \times w_{0}} \). These two kinds of latent representations are fused through a basic addition operation: \( \mathbf{z}_{r} = \mathbf{z}_{d} + \mathbf{z}_{s} \). The Image Retention block strategically integrates control signals from each layer into the primary U-Net, thereby facilitating image retention. The process for incorporating the control signal at the \( i \)-th layer is expressed as follows:
\begin{equation}
    \mathbf{z} =  F_{j}^{u}(\mathbf{z}, p) + R_{i}(\mathbf{z}, p, \mathbf{z}_{s})
\end{equation}
where $j = L - i + 1$, $L$ represents the total number of layers in the U-Net, and $p$ is the text embedding obtained by the CLIP text encoder.

\subsection{Classifier-free Guidance for Two Conditionings}\label{Classifier-free}

Classifier-Free Guidance~\cite{ho2022classifier} can balance the realism and diversity of generated images by adjusting the guidance weights. The implementation of Class-free guidance involves training a denoising model simultaneously on both conditional and unconditional inputs. During training, a simple approach is taken by probabilistically setting the condition to be a fixed null value $\varnothing$. During inference, both conditional and unconditional results are generated, and the guidance scale is utilized to control the distance between the generated results towards the conditional generation results and away from the unconditional generation results. Set the denoising network $ f_{\theta} $, condition $c$, and latent variable $z$ , the formula is as follows:
\begin{equation}
    \hat{f}_{\theta}(\mathbf{z}_{t}, c) = f_{\theta}(\mathbf{z}_{t}, \varnothing) + s \cdot (f_{\theta}(\mathbf{z}_{t}, c) - f_{\theta}(\mathbf{z}_{t}, \varnothing))
\end{equation}
In our model, we support two types of inputs: text $c_{t}$ and images $ c_{i} $. Hence, we explore the impact of classifier-free guidance on video generation under these two input conditions. During the training phase, the probabilities of setting text $c_{t}$ and images $c_{i}$ to empty text and blank white images, respectively, are both set at 10\%. Therefore, our model supports video generation in three scenarios: image-to-video, text-to-video, and image-text-to-video. The inference formulation during testing under these two control conditions is as follows:
\begin{equation}
\label{eq1}
\begin{aligned}
    \hat{f}_{\theta}(\mathbf{z}_{t}, c_{t}, c_{i}) & = f_{\theta}(\mathbf{z}_{t}, \varnothing, \varnothing) \\
    & + s_{t} \cdot (f_{\theta}(\mathbf{z}_{t}, c_{t}, \varnothing) - f_{\theta}(\mathbf{z}_{t}, \varnothing, \varnothing)) \\
    & + s_{i} \cdot (f_{\theta}(\mathbf{z}_{t}, c_{t}, c_{i}) - f_{\theta}(\mathbf{z}_{t}, c_{t}, \varnothing))
\end{aligned}
\end{equation}
Note \cref{eq1} is not the only possible result, and a detailed derivation of the formula can be found in the appendix. In \cref{fig: class-free}, we showcase the outcomes of the video generation conducted by the model under varying image guidance scales $s_{i}$. It can be observed that an increase in $s_{i}$ substantially intensifies the brightness and contrast of the videos generated. This indicates a trend within the generated results, progressively shifting towards the image domain.

\begin{figure}[!ht]
    \centering
    \includegraphics[width=.9\linewidth]{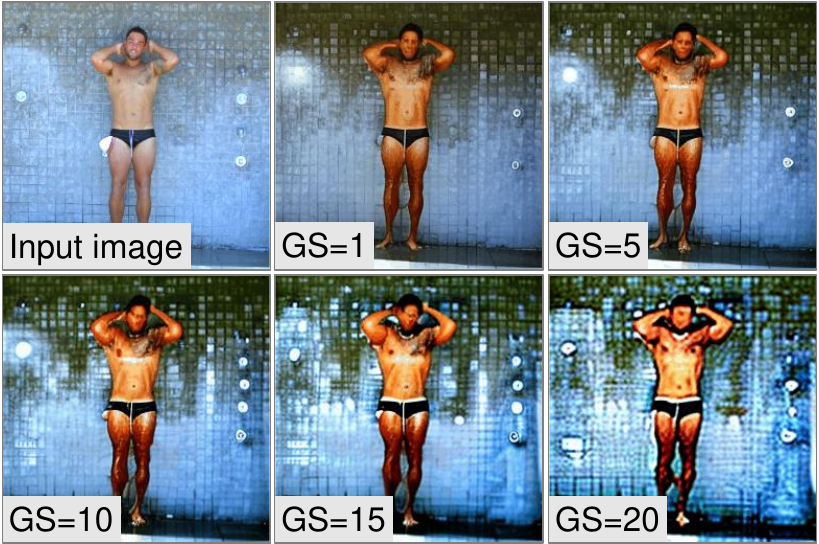}
    \caption{These are the generated first frames using different image guidance scales (GS) for classifier-free guidance.}
    \label{fig: class-free}
\end{figure}

\subsection{Inference}

During the inference phase, given the text and image input, DreamVideo can generate realistic and contextually consistent videos \( v \) that retain consistency between the given image and the initial frame of the generated video. The text is encoded by CLIP Text encoder~\cite{clip} to get text embedding $p$. Given Gaussian noise \( x_{0} \), the down-sample block (i.e., the Mid block, or \( F_{dm} \)) in U-Net produces a latent representation \( \mathbf{x}_{0}^{dm} = F_{dm}(p, \mathbf{x}_{0}) \). Concurrently, the Image Retention module \( R \) combines the Gaussian noise with the control signal from the image output, denoted as \( R(s, \mathbf{x}_{0}^{dm}) \). This aggregated information is then fed into the up-sample block \( F_{u} \) of U-Net, generating \( \mathbf{x}_{1}^{u} = F_{u}(p, \mathbf{x}_{0}^{dm}, R(s, x_{0}^{dm})) \). This generative process is repeated \( T \) times. Ultimately, the sampled latent code \( \mathbf{x}_{0} \) is decoded into pixel space by the decoder \( \mathcal{D} \).

We propose a Two-Stage Inference method, which uses the final frame of a previously generated video as the initial frame for the subsequent video. This strategy enables us to extend the video length by feeding the same text into the Two-Stage Inference process. Moreover, we can generate varied content videos by supplying different texts in two inference processes. Thanks to the great image retention capability of our DreamVideo, we can make the generated video longer and get generated video with two plots by the Two-Stage Inference method. The outcomes of our experiments using this method are presented in \cref{sec:two-stage}.

\section{Experiments}
\label{sec:exp}

\subsection{Experimental setup}


\textbf{Model architecture}
During the training phase, we loaded a pre-trained LVDMs model from ~\cite{gu2023reuse}. In our model, the trainable parameters are Temp layers in U-Net including Temp-Conv and Temp-Attn and the Image Retention Block, the number of trainable parameters is 680M.

\noindent \textbf{Training details} 
In our training, we utilize 340k data samples and trained for 5 epochs on a machine equipped with 8 GPUs, with a batch size of 9. Besides there is 10\% probability that images are replaced with white pictures, and 10\% probability that the text is empty. 
Our model can generate video in size $256 \times 256$ since the video generation resolution of pre-trained model is $ 256 $. In the training, we use $256 \times 256$ train data and the sample rate is $4$.

\noindent \textbf{Training datasets} 
Since the quality of the generated video depends on the quality of the datasets, we select the high-quality public dataset Pexels 300K~\cite{corran2022pexelvideos} for training data. This dataset contains videos with high resolution and strong diversity, particularly without any watermarks.

\begin{table*}[htbp]
    \centering
    \small
    \begin{tabular}{l c cccc cccc}
    \toprule
    \multirow{2}{*}{Methods} & \multirow{2}{*}{\#Videos} & \multicolumn{4}{c}{UCF101} & \multicolumn{4}{c}{MSR-VTT}     \\
    \cmidrule(lr){3-6} \cmidrule(lr){7-10}
    &  & FVD$\downarrow$ & IS$\uparrow$ &  FFF$_{\text{CLIP}}\uparrow$ & FFF$_{\text{SSIM}}\uparrow$  & FVD$\downarrow$ & IS$\uparrow$ & FFF$_{\text{CLIP}}\uparrow$ & FFF$_{\text{SSIM}}\uparrow$  \\
    \midrule
    I2VGen-XL & 10M & 526.94  & 18.90 & 0.68 & 0.17 & 341.72  & 10.52 & 0.57 & 0.17 \\
    VideoCrafter1 & 10.3M & 297.62 & 50.88 & \textbf{0.79} & 0.22  & 201.46  & 14.41 & \textbf{0.74} & 0.24   \\
    \midrule
    \rowcolor{danred} DreamVideo & 5.3M+340k~\tablefootnote{340k is the training data and 5.3 M is the training data of the pre-training model.} & \textbf{197.66}  & \textbf{54.39} & 0.76 & \textbf{0.37} & \textbf{149.18}  & \textbf{15.25} & \textbf{0.74} & \textbf{0.36} \\
    \bottomrule
    \end{tabular}
    \caption{Comparison with I2VGen-XL and VideoCrafter1 for zero-shot I2V generation on UCF101 and MSR-VTT.}
    \label{tab: main table}
\end{table*}%

\noindent \textbf{Evaluation}
The evaluation datasets are UCF101~\cite{soomro2012ucf101} with prompts sourced from ~\cite{gu2023reuse} and MSR-VTT~\cite{xu2016msr}. Following previous works~\cite{singer2022make,luo2023videofusion}, we employ four evaluation metrics:

\begin{enumerate}[label=(\roman*)]
  \item Fr\'{e}chet Video Distance (FVD)~\cite{unterthiner2019fvd}. Employing a trained I3D model~\cite{carreira2017quo} for FVD computation, as established in the Make-A-Video study~\cite{singer2022make}.
  \item Inception Score (IS)~\cite{saito2020train}. Following previous studies~\cite{singer2022make,hong2022cogvideo,luo2023videofusion}, we utilize a trained C3D model to compute the video version of the IS.
  \item CLIP Similarity (CLIP-SIM). We compute the CLIP~\cite{radford2021learning} text-image similarity for each individual frame and calculate the average score for the text prompt. Our evaluation utilizes the ViT-B/32 CLIP model as its backbone.
  \item First-Frame Fidelity by CLIP (FFF$_{\text{CLIP}}$) measures the similarity between the initial frames of real and generated videos by CLIP visual encoder.
  \item First-Frame Fidelity by SSIM~\cite{SSIM} (FFF$_{\text{SSIM}}$) assesses the similarity between two images based on luminance, contrast, and structural integrity. The FFF$_{\text{SSIM}}$ ranges from 0 to 1, where higher values denote lower image distortion.
\end{enumerate}

\subsection{Qualitative evaluation}
\begin{figure}[!ht]
    \centering
    \includegraphics[width=1.0\linewidth]{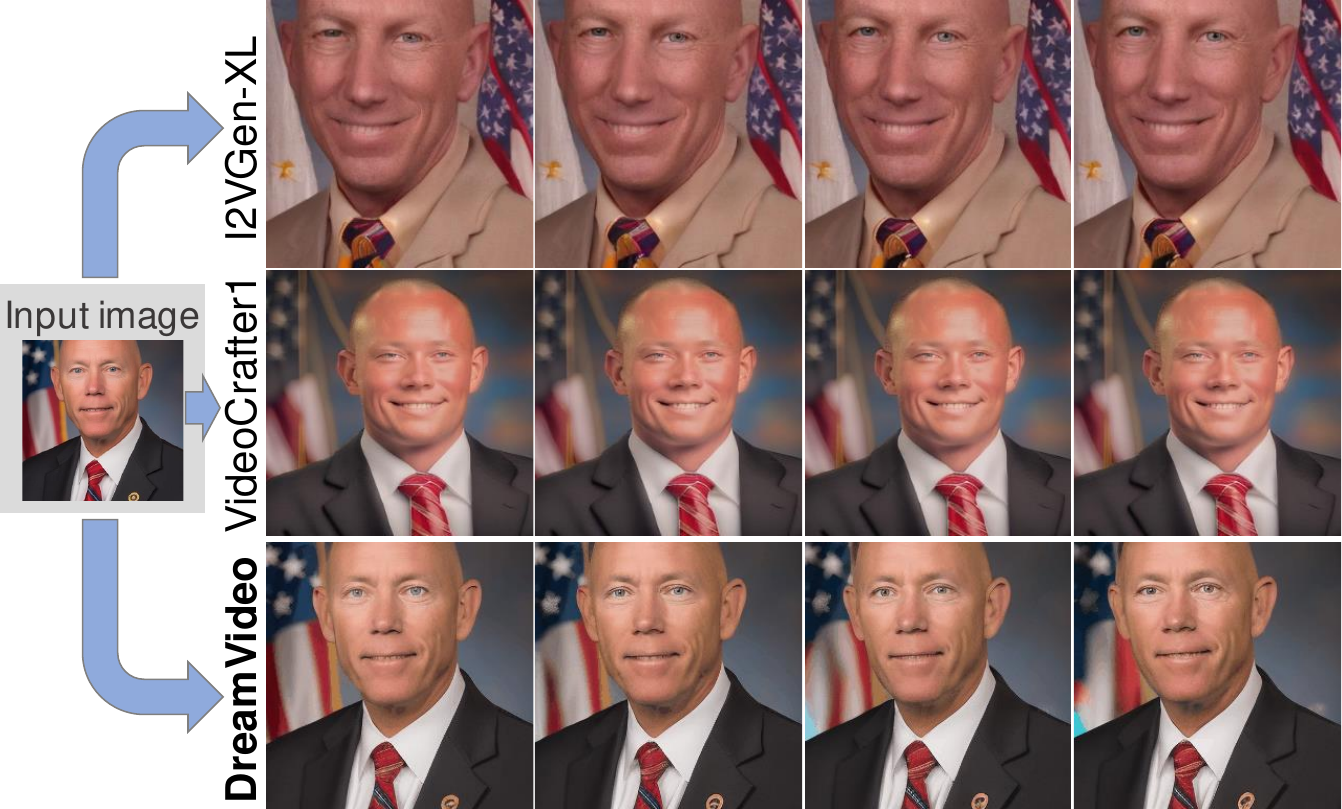}
    \caption{Comparative analysis across methods: I2Vgen-XL, VideoCrafter1, and Ours (top to bottom arrangement). Leftmost is the original image, columns 1-3 indicate frames 0-3, and the final column presents the last frame.}
    \label{fig:vis1}
\end{figure}

We conduct a qualitative comparison of our approach against recent advancements, especially I2VGen-XL~\cite{zhang2023i2vgen} and VideoCrafter1~\cite{chen2023videocrafter1}. VideoCrafter1 is responsible for the synthesis of videos from textual and image inputs while I2VGen-XL is uniquely oriented towards image-to-video generation. \cref{fig:vis1} provides an illustrative comparison of the methods being evaluated. The third row presents the outcomes of our approach, whereas the first and second rows demonstrate the outputs of I2VGen-XL and VideoCraft1 respectively. We found that the videos produced utilizing our method closely resemble the source images. On the other hand, the results from I2VGen-XL show significant variations including changes in character identities and discrepancies in color and positional representations. VideoCrafter1, while superior to I2VGen-XL in terms of adhering to the original images, also shows noticeable deviations. Further analysis of the initial and eighth frames created by these models revealed minimal perceptible motion in both I2VGen-XL and VideoCrafter1 outputs, exhibiting an almost static state. In stark contrast, the videos generated by our method display noticeable motion fluctuations, highlighting its superior dynamism.

\begin{figure*}[!ht]
    \centering
    \includegraphics[width=0.75\linewidth]{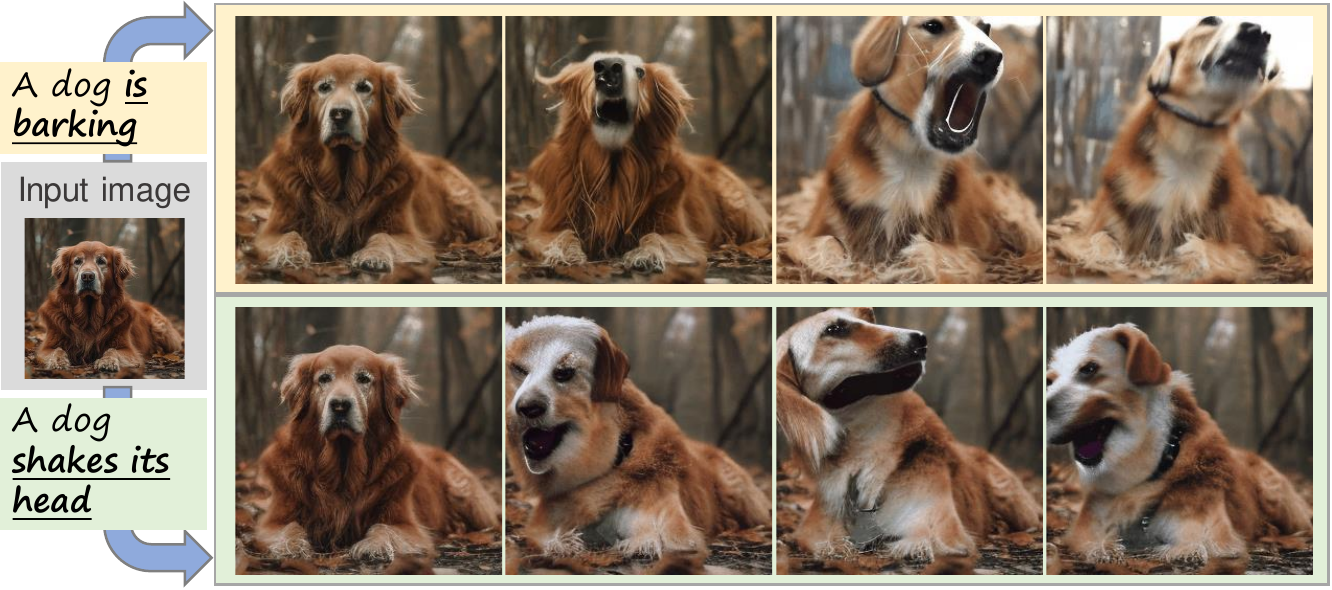}
    \caption{Given the same image, different text prompts can lead to different output videos. This can be seen as evidence of our text-guidance capability for controlling video content motion.}
    \label{fig: act}
\end{figure*}

\begin{figure}[!ht]
    \centering
    \includegraphics[width=1.0\linewidth]{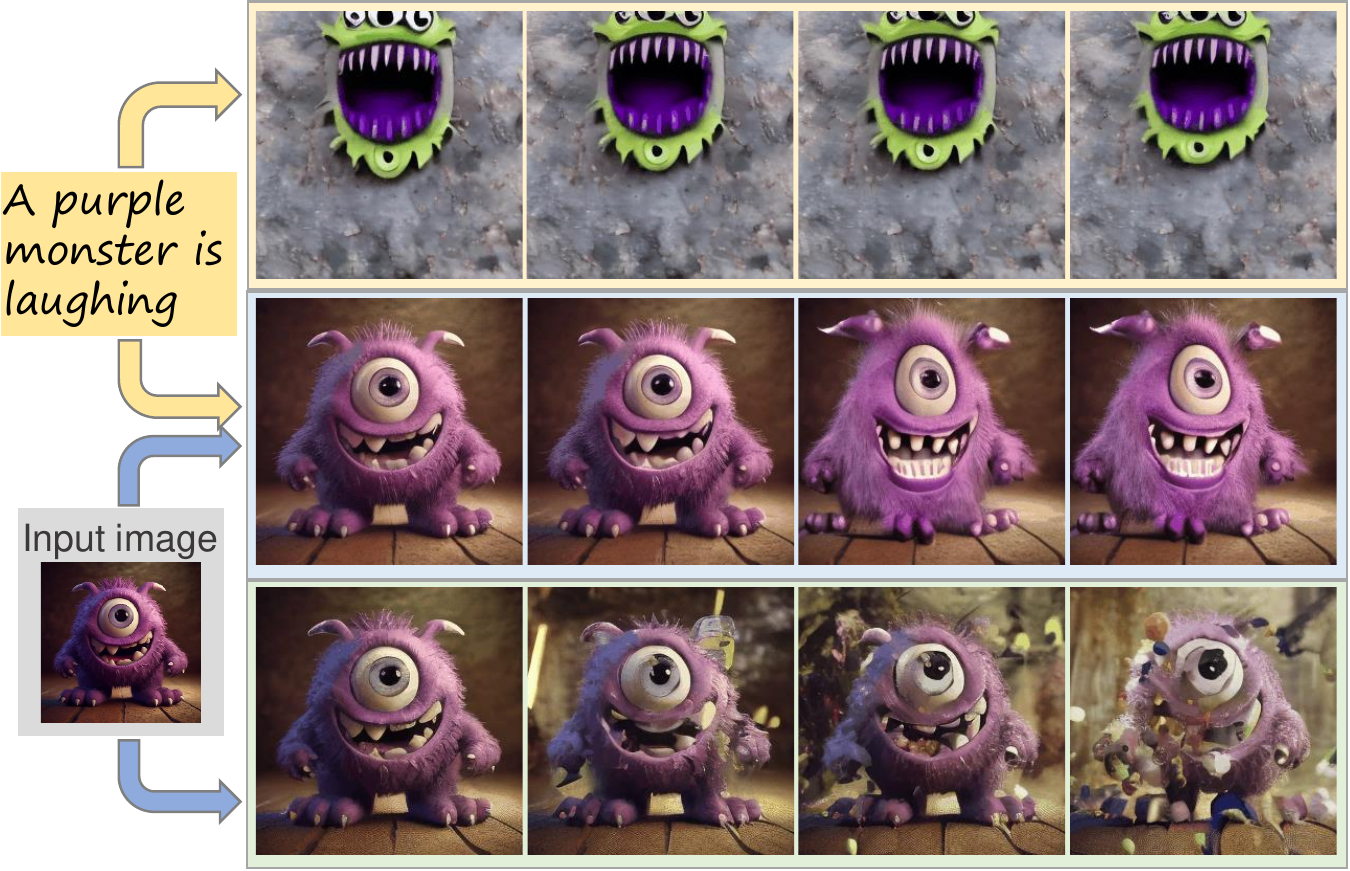}
    \caption{This is an ablation on using one or both of an input image and a text prompt. We can find that using both can lead to the best result. Using only text prompt leads to low fidelity and using only image leads to poor temporal consistency.}
    \label{fig: multi}
\end{figure}

\subsection{Quantitative evaluation}\label{Quantitative evaluation}

\textbf{Comparison to other models}
We carry out quantitative assessments on the UCF101~\cite{soomro2012ucf101} and MSR-VTT~\cite{xu2016msr} datasets, and compare our model outcomes to the models I2VGen-XL and VideoCrafter1. As evidenced by \cref{tab: main table}, our model consistently presents the lowest Fréchet Video Distance (FVD) scores on both datasets. This proves that our model can generate video with better continuity.
Regarding the Inception Score (IS), our model's IS score is higher than that of other methods, which means that the videos generated by DreamVideo have the best quality.
As for the $FFF_{CLIP}$ metric, this is derived by employing clips to calculate the similarity between the initial frames of the generated and original videos. However, we believe that this metric may not fully represent the quality of the generated initial frame. The visual features encoded in the clip images are coarse and focus more on the presence of objects in the image rather than the quality of the initial frame. 
FFF$_{\text{SSIM}}$ evaluates the similarity between the initial frame of the generated video and a given image in terms of brightness, contrast, and structural integrity. As evident from \cref{tab: main table}, our model significantly outperforms other models in terms of FFF$_{\text{SSIM}}$, which demonstrates that our model's ability to generate video with high image retention.

\noindent \textbf{Compare to T2V and TI2V}
Our evaluation of the model in both Text2Video (T2V) and TextImage2Video (TI2V) contexts is conducted on the UCF101 and MSR-VTT datasets. As indicated in \cref{tab: ours}, our model exhibits a notable performance advantage in textimage2video over text2video on both datasets, especially in the FVD and IS metrics. This enhanced performance aligns with expectations, as the model leverages the first frame of the original video, which contributes to higher FVD and IS scores. However, a peculiar observation on the MSR-VTT dataset is the marginally lower CLIP-SIM score for textimage2video compared to text2video. This variation is attributed to the frequent misalignment between the text and initial frames within the MSR-VTT dataset. The formidable initial frame retention ability of our model often leads to a mismatch between the generated video and the accompanying caption in the textimage2video scenario. Consequently, this incongruence is reflected in the lower CLIP-SIM scores observed for our model on the MSR-VTT dataset in the textimage2video condition.

\begin{table}[htbp]
    \centering
    \small
    \begin{tabular}{l cc ccc}
    \toprule
    \multirow{2}{*}{Methods} & \multicolumn{2}{c}{UCF101} & \multicolumn{3}{c}{MSR-VTT}     \\
    \cmidrule(lr){2-3} \cmidrule(lr){4-6}
     & FVD↓ & IS↑  & FVD↓ & IS↑ & CLIP-SIM↑   \\
    \midrule
    T2V & 463.76  & 22.82  & 250.80  & 10.90   & \textbf{20.97}   \\
    TI2V & \textbf{214.52}  & \textbf{46.31} & \textbf{167.80}  & \textbf{13.43}     & 20.55    \\
    \bottomrule
    \end{tabular}
    \caption{Comparison of our model Text2Video (T2V) and TextImage2Video (TI2V) generation on UCF101 and MSR-VTT.}
    \label{tab: ours}
\end{table}%

\begin{figure*}[!ht]
    \centering
    \includegraphics[width=1.0\linewidth]{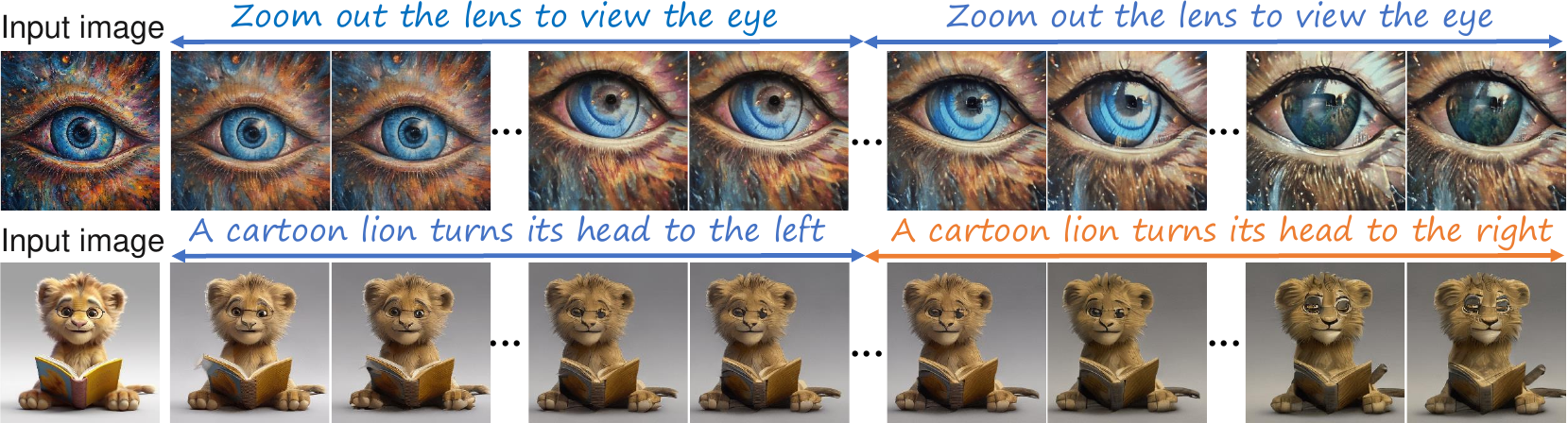}
    \caption{Outcomes of Two-Stage Inference: same text input in the first row, different texts in the second.}
    \label{fig:two}
\end{figure*}

\subsection{Varied textual inputs}

In our experiments, we explored the generation of videos from a single initial frame using varied textual inputs. The first row of \cref{fig: act} presents results from the input text \textit{``A dog is barking''} resulting in a corresponding video. Conversely, as shown in the second row of the same figure, inputting a distinct action text, \textit{``A dog shakes its head''}, leads to a markedly different video outcome compared to the first. This demonstrates DreamVideo's versatility in producing diverse videos from any given initial frame, contingent on the provided text. A notable aspect of DreamVideo is its remarkable image retention capability, evident in the consistent quality of the output videos despite variations in the textual input while maintaining the same initial frame.

\subsection{Two-Stage inference}\label{sec:two-stage}
Our model demonstrates a significant capacity for maintaining the initial frame in video outputs when given an initial control image, which allows for the exploration of a two-stage inference process. This process entails using the output video from the first stage as the initial image input for generating a video with a preserved initial frame. Employing this two-stage inference enables our model to produce videos with up to 16 frames, maintaining continuity owing to its robust image retention capabilities.
As depicted in the first row of \cref{fig:two}, utilizing the same textual input across both stages results in an extended video length. Furthermore, as shown in the second row of \cref{fig:two}, when different textual inputs are applied in each stage of inference, our model is capable of generating varied actions within a single video. Remarkably, despite these variations, the video retains consistency and a seamless flow.

\subsection{Multi-Combinations of video generation}

We investigate the output of the model by utilizing various combinations of inputs to test the model's video generation capability under different conditions. As depicted in the first row of \cref{fig: multi}, when the model is guided solely by text to generate the video, it tends to produce a video of lower quality. When we provide the model with an initial frame but without any text, the content of the generated video is quite ambiguous. However, as presented in the last row of \cref{fig: multi}, when we provide both the initial frame and guiding text, the model can generate a video in which the content is clearly defined by the text. Besides the generated video is of a high quality and the initial frame is very well preserved.

\subsection{User Study}

We conduct a user study to assess our DreamVideo, I2VGen-XL, and VideoCrafter1 across three key metrics: Initial Frame Retention, Prompt Alignment, and Video Quality. The study involves thirty unique text-image-video combinations, and ten participants are tasked with rating the output produced by each of the three models across these metrics. We use Likert scale~\cite{likert1932technique} surveys to analyze quantitative data, where users have the option to select from five rating levels, ranging from 1 (Extremely Dissatisfied) to 5 (Extremely Satisfied).
As depicted in \cref{Quantitative evaluation}, we find our DreamVideo achieves the highest score in fidelity and VideoCrafter1 follows closely, while I2VGen-XL lags behind. This outcome corroborates the conclusion drawn in our previous \cref{tab:use study}. 
In the context of text-video alignment, all three models displayed comparable performances. 
In terms of video quality metrics, our model outperforms I2VGen-XL but slightly trails behind VideoCrafter1. This discrepancy arises due to the resolution disparities in videos generated by VideoCrafter1 ($320\times512$) and our model ($256\times256$).

\begin{table}[htbp]
    \centering
    \small
    \setlength{\tabcolsep}{5pt}
    \begin{tabular}{lccc}
    \toprule
    Methods & Fidelity & Alignment & Quality \\
    \midrule
    I2VGen-XL~\cite{zhang2023i2vgen} & 3.64  & 4.04  & 3.80   \\
    VideoCrafter1~\cite{chen2023videocrafter1} & 3.95  & 4.23  & \textbf{3.92}      \\
    \midrule
    \rowcolor{danred}  DreamVideo (Ours) & \textbf{4.17}  & \textbf{4.28} & 3.88  \\
    \bottomrule
    \end{tabular}
    \caption{The results of our user study. \textit{Fidelity} represents the visual similarity between the input image and the output video. \textit{Alignment} is for evaluating text-video consistency. \textit{Quality} is the visual perception of the output video.}
    \label{tab:use study}
\end{table}%


\section{Conclusion}
\label{sec:concl}
In this work, we present DreamVideo, a model for synthesizing high-quality videos from images. Our DreamVideo has a great image retention capability and supports a combination of image and text inputs as controlling parameters.
We propose an Image Retention block that combines control information and gradually integrates it into the primary U-Net. We Explore double-condition class-free guidance for different degrees of image retention. It's noteworthy that one limitation of our model is that the image retention ability of our DreamVideo relies on high-quality training data. Our DreamVideo is enabled to generate video that maintains superior image retention quality through training with high quality datasets. Finally, we demonstrate DreamVideo’s superiority over the open-source image-video model qualitatively and quantitatively.

{
    \small
    \bibliographystyle{ieeenat_fullname}
    \bibliography{main}
}


\end{document}



\clearpage
\setcounter{page}{1}
\maketitlesupplementary

\section{Classifier-free guidance formula}
\label{sec:Classifier-free formula}

In the presence of both textual condition $c_{t}$ and visual condition $c_{i}$, the video generation model can be formulated as $ F_(\mathbf{z} | c_{t}, c_{i}) $, where $\mathbf{z}$ represents the latent variable. By applying Bayes' theorem, we obtain the following formula:
\begin{equation*}\label{eq:bes}
    F(\mathbf{z}|c_{t},c_{i}) = \frac{F(\mathbf{z},c_{t},c_{i})}{F(c_{t},c_{i})} = \frac{F(c_{i} | c_{t},\mathbf{z})F(c_{t}|\mathbf{z}) F(\mathbf{z})}{F(c_{t},c_{i})}
\end{equation*}
For the convenience of differentiation, we take the logarithm of both sides of the above equation:
\begin{equation*}\label{eq:log}
\begin{aligned}
    \log(F(\mathbf{z}|c_{t},c_{i})) = & \log(F(c_{i} | c_{t}, \mathbf{z})) + \log(F(c_{t}|\mathbf{z})) \\
    + & \log(F(\mathbf{z})) - \log(F(c_{t},c_{i}))
\end{aligned}
\end{equation*}
Taking the derivative of both sides of the equation yields the following:
\begin{equation*}\label{eq:de}
\begin{aligned}
    \nabla \log(F(\mathbf{z}|c_{t},c_{i}))  = & \nabla \log(F(c_{i} | c_{t},\mathbf{z}))  \\
     + & \nabla \log(F(c_{t}|\mathbf{z})) +  \nabla \log(F(\mathbf{z}))
\end{aligned}
\end{equation*}
The terms are corresponding to Eq.(\textcolor{red}{5}) in our paper, where variable $s_{t}$ plays a pivotal role in guiding text-influenced synthesis of videos and $s_{i}$ regulates the degree of image retention during the video synthesis process. Furthermore, our analysis indicates that variations in the decomposition order within $F(\mathbf{z},c_{t},c_{i})$ result in the derivation of distinct formulas. In situations involving multiple decompositions, this leads to a spectrum of formulaic outcomes. We find the specific formula presented in our study more effectively illustrates video generation under dual conditions in a classifier-free guidance framework.

\section{Human evaluations details}
\label{sec:Human}
We perform a manual evaluation on a corpus of 30 videos, which are produced using three distinct models: DreamVideo, VideoCrafter1, and I2VGen-XL. This evaluation is based on ten unique sets of image-text pairs. A group of 10 volunteers, selected at random, conduct the evaluation, focusing on the following three criteria:
\begin{itemize}
    \item Fidelity: This criterion assesses the degree of similarity between the initial frame of the generated video and the supplied image.
    \item Alignment: This evaluates the congruence of the generated video with the corresponding text description.
    \item  Quality: This overarching criterion evaluates the overall standard of the generated videos, incorporating the volunteers' comprehensive impressions.
\end{itemize}
Each aspect is meticulously analyzed to ensure a comprehensive assessment of the video generation capabilities of the models with both text and image inputs.

\section{Qualitative evaluation in more data}

We utilize the entire Pexels 300K dataset (about 340K videos) and train 5 epochs on 16 GPUs with the same hyperparameters as in the paper. We conduct same quantitative experiments on UCF101 and MSRVTT, as shown in \cref{tab: appendix}. The experiments demonstrate that our DreamVideo outperforms other methods in FVD and IS metrics, achieving state-of-the-art results in the IT2V task, even with a small amount of data.

\begin{table}[htbp]
    \centering
    \small
    \setlength{\tabcolsep}{4pt}
    \begin{tabular}{l c cc cc}
    \toprule
    \multirow{2}{*}{Methods} & \multirow{2}{*}{\#Videos} & \multicolumn{2}{c}{UCF101} & \multicolumn{2}{c}{MSR-VTT}     \\
    \cmidrule(lr){3-4} \cmidrule(lr){5-6}
    &  & FVD↓ & IS↑  & FVD↓ & IS↑  \\
    \midrule
    I2VGen-XL & 10M & 526.94  & 18.90  & 341.72  & 10.52 \\
    VideoCrafter1 & 10.3M & 297.62  & 50.88 & 201.46  & 14.41   \\
    \midrule
    DreamVideo-S & 5.3M+188K  & 214.52 & 46.31  & 167.80 & 13.43   \\
    \rowcolor{danred} DreamVideo-L & 5.3M+340k & \textbf{197.66}  & \textbf{54.39} & \textbf{149.18}  & \textbf{15.25}   \\
    \bottomrule
    \end{tabular}
    \caption{Comparison with I2VGen-XL and VideoCrafter1 for zero-shot I2V generation on UCF101 and MSR-VTT.}
    \label{tab: appendix}
\end{table}%
%
\begin{figure*}[!ht]
    \centering
    \includegraphics[width=1.0\linewidth]{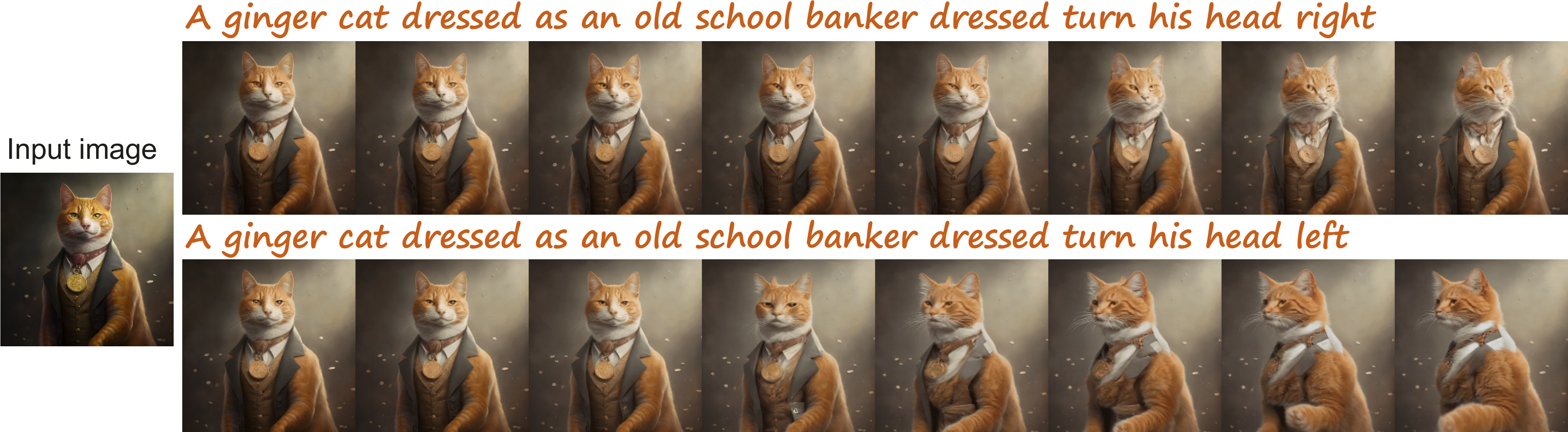}
    \caption{Videos with varied textual inputs in $512 \times 512$.}
    \label{fig:i2v-varied-512}
\end{figure*}
\begin{figure*}[!ht]
    \centering
    \includegraphics[width=1.0\linewidth]{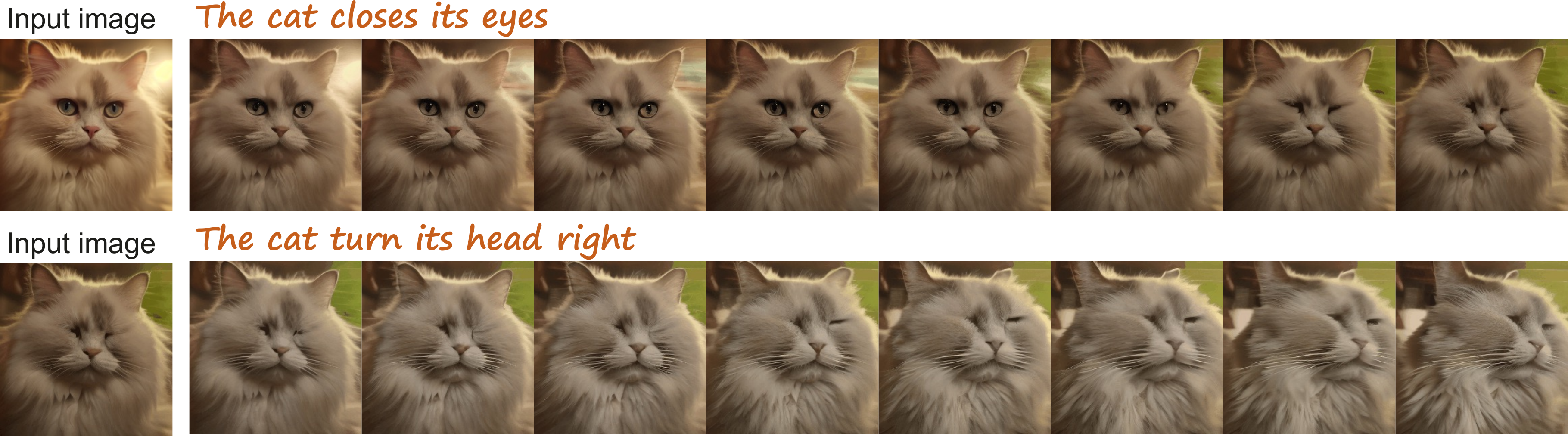}
    \caption{Videos of two-stage inference in $256 \times 256$.The initial frame of video in the second stage (second row) is the same as the final frame of the first stage (first row).}
    \label{fig:i2v-twostage}
\end{figure*}
\section{More experimental results}\label{sec:more-experimental-results}
This section showcases a broader array of DreamVideo examples, further illustrating the model's image retention capabilities. All images used for video generation are from Midjourney. More examples in video format can be found in the project homepage \url{https://sense39.github.io/DreamVideo/}.

\subsection{More video examples}

Building upon DreamVideo's default configuration, we present an extended set of experimental examples at a resolution of $256 \times 256$, as detailed below:

\subsubsection{Image-to-video examples}\label{supp:Image-to-Video}
\begin{figure*}[!ht]
    \centering
    \includegraphics[width=1.0\linewidth]{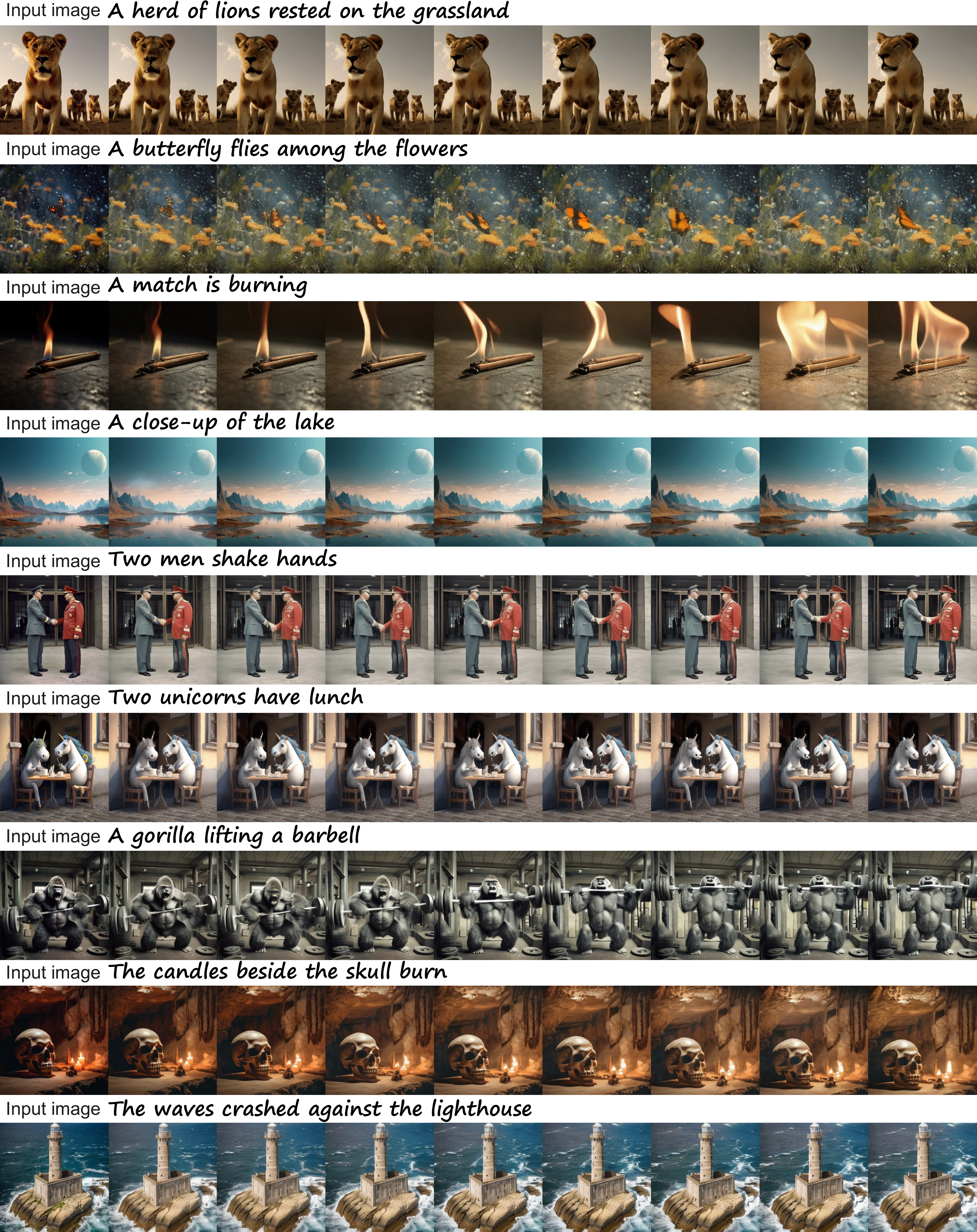}
    \caption{Image-to-video results in $256 \times 256$.}
    \label{fig:i2v-256}
\end{figure*}

This subsection shows additional image-to-video results, highlighting our model's robust image retention capability at $256 \times 256$ resolution (\cref{fig:i2v-256}).

\subsubsection{Two-Stage inference}\label{supp:two-stage}
Additional examples of two-stage video generation are provided, showcasing the model's two-stage inference capabilities in \cref{fig:i2v-twostage}.

\subsubsection{Varied textual inputs inference}\label{supp:two-stage}
\begin{figure*}[!ht]
    \centering
    \includegraphics[width=1.0\linewidth]{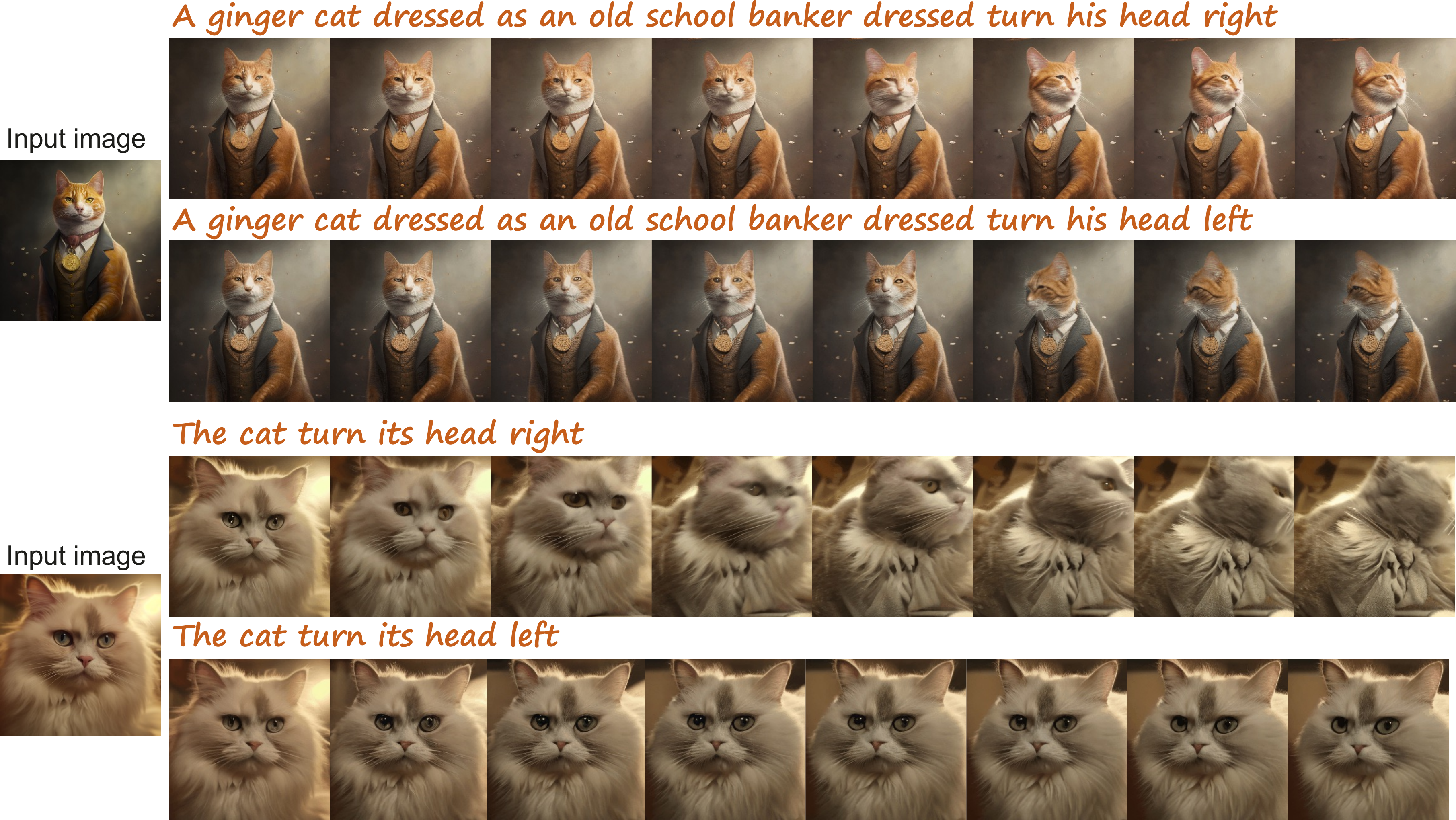}
    \caption{Videos with varied textual inputs in $256 \times 256$.}
    \label{fig:i2v-varied}
\end{figure*}

We exhibit more examples of video generation with varied textual inputs at $ 256 \times 256 $ resolution in \cref{fig:i2v-varied}.

\subsection{More video at high resolution}
In the standard configuration, our DreamVideo model is capable of generating videos at a resolution of $256 \times 256$, leveraging both image and text inputs for control. Upon further exploration, we find that without the need for additional training, a direct modification of the original model's settings to a resolution of $512$ facilitates the production of videos at $512 \times 512$ resolution.
However, this method introduces a significant limitation: the videos generated exhibit considerable instability, which is most notably manifested as a distinct discontinuity between the first and second frames.

\subsubsection{Image-to-video examples}\label{supp:Image-to-Video}

\begin{figure*}[!ht]
    \centering
    \includegraphics[width=1.0\linewidth]{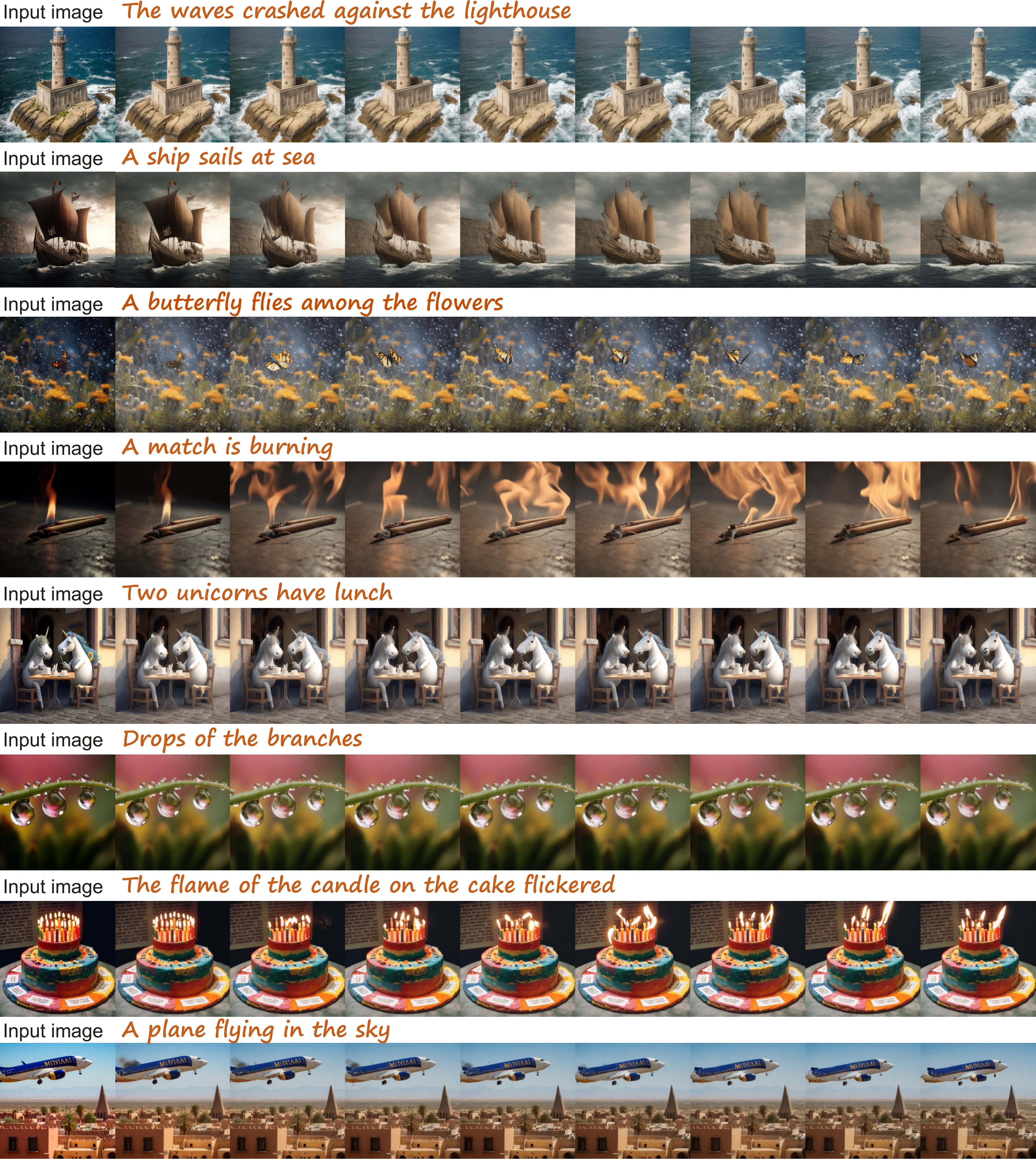}
    \caption{Image-to-video results in $512 \times 512$.}
    \label{fig:i2v-512}
\end{figure*}

There are image-to-video examples in $512 \times 512 $ as shown in \cref{fig:i2v-512}.

\subsubsection{Varied textual inputs inference}

There are examples under the varied text control in $512 \times 512 $ as shown in \cref{fig:i2v-varied-512}.

\section{Limitations}
The efficacy of DreamVideo is primarily constrained by two factors: the quality and quantity of the datasets employed. The dataset quality is pivotal in ensuring the model's proficiency in sustaining high image fidelity. Concurrently, the dataset quantity significantly influences the model's robustness across diverse scene representations.
Additionally, computational constraints necessitated a compromise in the training process. To reduce computational demands, the resolution of videos generated by our model is limited to $256 \times 256$ pixels. This restriction, while computationally efficient, results in a loss of detail in the produced videos. 
However, our model possesses the capability to generate videos at a higher resolution of $512 \times 512$ pixels without retraining, which markedly improves detail richness. It is noteworthy that this enhancement in high resolution, achieved without retraining, introduces certain challenges. It leads to a lack of continuity between the first and second frames of the generated videos, alongside overall instability in the video generation process. Future research endeavors will be directed towards resolving this dichotomy: achieving high-resolution video generation at $512 \times 512$ pixels while ensuring stability, without necessitating additional training.
